%% file: main.tex
\documentclass[10pt,twocolumn,letterpaper]{article}

\usepackage[pagenumbers]{iccv} % To force page numbers, e.g. for an arXiv version
\definecolor{iccvblue}{rgb}{0.21,0.49,0.74}
\usepackage[pagebackref,breaklinks,colorlinks,allcolors=iccvblue]{hyperref}

\title{Teaching Humans Subtle Differences with \textit{DIFF}usion }

\author{Mia Chiquier$^{*}$ \\ Columbia University \\ {\tt\small mia.chiquier@cs.columbia.edu}
\and Orr Avrech$^{*}$ \\ Columbia University \\ {\tt\small oa2429@columbia.edu}
\and Yossi Gandelsman \\ UC Berkeley \\ {\tt\small yossi\_gandelsman@berkeley.edu} \\
\and Berthy Feng \\ California Institute of Technology \\ {\tt\small bfeng@caltech.edu}
\and Katherine Bouman \\ California Institute of Technology \\ {\tt\small klbouman@caltech.edu}
\and Carl Vondrick \\ Columbia University \\ {\tt\small vondrick@cs.columbia.edu}
}

\begin{document}

\twocolumn[{%
        \renewcommand\twocolumn[1][]{#1}%
        \maketitle
        \begin{center}
        \captionsetup{type=figure}
                \vspace{-0.5cm}
\includegraphics[width=1.0\linewidth]{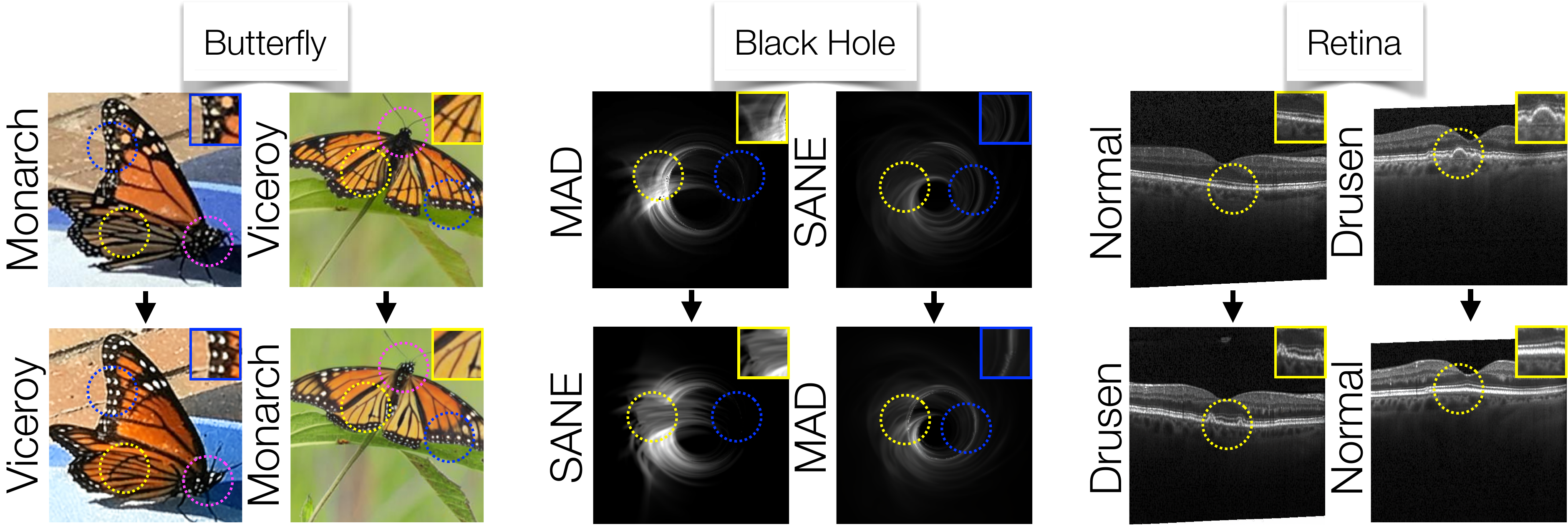}
        \captionof{figure}{\textbf{\textit{DIFF}usion Counterfactuals.} We illustrate the counterfactual results from our methods on the Butterfly dataset, the Black Hole dataset, and the Retina dataset. In the Butterfly dataset, the Viceroy has a cross-sectional line (\textbf{\color{yellow}{yellow}}), a smaller head with less dots (\textbf{\color{magenta}{magenta}}), and more ``scaley'' dots (\textbf{\color{blue}{blue}}), compared to the Monarch. In the Black Hole dataset, SANE has more uniform wisps (\textbf{\color{yellow}{yellow}}) and less of a prominent photon ring (\textbf{\color{blue}{blue}}) as compared to MAD, with these distinguishing features discovered through our method rather than known a priori. In the Retina dataset, normal retinas lack the horizontal line bumps (\textbf{\color{yellow}{yellow}}) present in retinas with drusen.}{\vspace{0.1cm}
}
               
                \label{fig:teaser}
        \end{center}
}]

% Place abstract and intro in the same column by removing the \input command
% and directly including the abstract content here
% \input{/sec/0_abstract} 
% \input{sec/1_intro}
% \input{sec/2_relatedwork}
% \input{sec/3_method}
% \input{sec/4_experiments}
% \input{sec/5_discussion}

% arxiv version
% \begingroup
% \let\thefootnote\relax
% \footnotetext{$^{*}$Equal contribution.}
% \endgroup
\input{sec_arxiv/0_abstract} 
% \let\thefootnote\relax
% \footnotetext{$^{*}$Equal contribution.}
\begingroup
\let\thefootnote\relax
\footnotetext{$^{*}$Equal contribution.}
\endgroup

% Reset footnote counter for regular footnotes in the document
\setcounter{footnote}{0}

\input{sec_arxiv/1_intro}
\input{sec_arxiv/2_relatedwork}
\input{sec_arxiv/3_method}

\input{sec_arxiv/4_experiments}

\input{sec_arxiv/5_discussion}
{
    \small
    \bibliographystyle{ieeenat_fullname}
    \bibliography{main}
}

 \input{sec_arxiv/X_suppl}

\end{document}

%% file: sec_arxiv/0_abstract.tex
\begin{abstract}
Scientific expertise often requires recognizing subtle visual differences that remain challenging to articulate even for domain experts. We present a system that leverages generative models to automatically discover and visualize minimal discriminative features between categories while preserving instance identity. Our method generates counterfactual visualizations with subtle, targeted transformations between classes, performing well even in domains where data is sparse, examples are unpaired, and category boundaries resist verbal description. Experiments across six domains, including black hole simulations, butterfly taxonomy, and medical imaging, demonstrate accurate transitions with limited training data, highlighting both established discriminative features and novel subtle distinctions that measurably improved category differentiation. User studies confirm our generated counterfactuals significantly outperform traditional approaches in teaching humans to correctly differentiate between fine-grained classes, showing the potential of generative models to advance visual learning and scientific research.
%Human expertise depends on the ability to recognize subtle visual differences, such as distinguishing diseases, species, or celestial phenomena. We propose a new method to teach novices how to differentiate between nuanced categories in specialized domains. Our method uses generative models to visualize the minimal change in features to transition between classes, i.e., counterfactuals, and performs well even in domains where data is sparse, examples are unpaired, and category boundaries are not easily explained by text. By manipulating the conditioning space of diffusion models, our proposed method \textit{DIFF}usion disentangles category structure from instance identity, enabling high-fidelity synthesis even in challenging domains. Experiments across six domains show accurate transitions even with limited and unpaired examples across categories. User studies confirm that our generated counterfactuals outperform unpaired examples in teaching perceptual expertise, showing the potential of generative models for specialized visual learning.
\end{abstract}

%% file: sec_arxiv/1_intro.tex
\section{Introduction}
\label{sec:intro}

Generative models, especially large-scale image diffusion models, have transformed text-to-image creation, opening new ways to visualize concepts across various domains. While these models excel in everyday contexts with clear category distinctions, a far more challenging frontier exists in scientific fields where visual differences between categories are so subtle that they often remain unknown and unidentified even to domain experts.

In specialized scientific domains, the complete set of visual features distinguishing between categories may be partially or entirely undiscovered. For example, astronomers studying black hole simulations have no established verbal characteristics to differentiate MAD from SANE models because these distinguishing features have not yet been comprehensively identified. Entomologists may differentiate Viceroy and Monarch butterflies through the Viceroy’s characteristic cross-sectional black line, yet may miss other distinguishing features that could further help the differentiation. This represents the fundamental challenge for visual expertise training: how do we teach recognition of patterns we ourselves don’t fully understand?

  One of the most effective ways to reveal subtle category differences is to transform an image and rapidly flip between the original and its altered version. In scientific domains, this approach faces three key challenges: (1) automatically identifying discriminative features that may not be known or easily articulated even by experts, (2) limiting changes exclusively to these category-defining features, and (3) preserving all other identity characteristics of the instance. We develop a system that combines state-of-the-art image editing techniques with visual algebraic conditioning guidance to address these challenges in data-scarce scientific domains. Our approach automatically identifies discriminative features through visual algebraic operations that extract category-specific information without requiring explicit articulation. By integrating inverted noise maps (z) to preserve identity features with conditioning vectors (c) that guide category transformations, our system achieves effective identity-preserving yet category-changing results, that isolate and visualize subtle differences between scientific categories.

% We develop a simple but effective way to condition diffusion models to make these precise, minimal changes, thus revealing sublte differences that might otherwise remain undetected. 

% We develop a system that combines state-of-the-art image editing techniques with visual algebraic conditioning guidance to create image transformations in data-scarce scientific domains. By integrating inverted noise maps (z) to capture identity features with conditioning vectors (c) that guide category transformations, our system achieves more identity-preserving yet in-distribution category-changing results, as compared to previous methods. 
% Traditional approaches rely on experts explaining differences to students, but these explanations are inherently limited by what experts have consciously identified. We cannot teach what we do not yet know. Side-by-side comparisons help but fail to isolate specific features from confounding variables. The ideal solution would generate transformations that highlight precisely what changes between categories, potentially revealing patterns that remain to be discovered.

Our approach overcomes limitations in current counterfactual visualization methods, which have traditionally been applied in domains where category distinctions are already well-understood and easily verbalized. Text-guided editing methods rely on linguistic descriptions, which can be too ambiguous to specify desired visual changes. Methods like Concept Sliders' ~\cite{gandikota2023conceptslidersloraadaptors} effectiveness, which is guided by the image distributions themselves, depend on paired examples in most cases—a constraint limiting their use in teaching scenarios. Visual counterfactual generation methods often rely on gradients from a classifier, a limitation when data is scarce. Classifier-free alternatives, like TIME~\cite{Jeanneret_2024_WACV}, struggle with image quality and coherence for subtle differences.

% Traditional methods struggle in specialized scientific contexts where the distinguishing features themselves are difficult to verbalize or remain undiscovered, paired data is unavailable, and limited data makes classifier guidance based methods impractical. Our method makes counterfactual visualization undven clar these challenging conditions possible. 
% We propose a system tailored for subtle differences, carefully assembling  \cv{``assembling'' reads incremental} image-editing techniques specifically for their ability to capture and highlight nuanced visual distinctions. Our approach proposes a simple algebraic technique \cv{algebraic technique is not key to highlight  - the more important point is how you combine the low-level with the high-level guidance in z and c.} for conditioning that enables transformations preserving instance-specific features while emphasizing the distinctive characteristics between categories in data-scarce scientific contexts.

Through experiments across six domains, we demonstrate our approach’s effectiveness in highlighting visual differences between categories. For instance, in black hole simulations, where distinguishing characteristics between MAD and SANE models remain largely unknown, our counterfactual visualizations emphasize distinct visual patterns in the image distribution. The transformations draw attention to variations in the uniformity of wisps and prominence of the photon ring, which are features that black hole experts themselves had not identified. 

User studies confirm the effectiveness of our approach: participants who trained with our counterfactual visualizations demonstrated significantly better category differentiation performance than those using traditional approaches with unpaired images. This validates that our method highlights meaningful visual patterns that can be used to build expertise, even when those subtle patterns have not yet been explicitly identified or understood.

%% file: sec_arxiv/2_relatedwork.tex
% \begin{figure*}[t]
%   \centering
%   \includegraphics[width=1.0\linewidth]{figures/method_diagram_v4_horiz_helve.pdf} 

%    \caption{\textbf{\textit{DIFF}usion method overview.} }
%    \label{fig:method-diagram}
% \end{figure*}
% \begin{figure*}[t]
%   \centering
%   \includegraphics[width=\linewidth]{figures/method_clip.jpeg} 

%    \caption{\textbf{\textit{DIFF}usion method overview.} }
%    \label{fig:method-diagram}
% \end{figure*}

\section{Related Work}
\label{sec:relatedwork}

\paragraph{Visual Counterfactual Explanations.} A counterfactual image shows how an input would appear if altered to switch its class, enhancing interpretability. Counterfactual inference crafts images that not only differ in classification but also clarify the visual features defining each distribution. Approaches for visual counterfactual explanations (VCEs) make use of generative model edits, with VAEs \cite{rodriguez2021trivialcounterfactualexplanationsdiverse}, GANs \cite{lang2021explainingstyletraininggan}, and more recently, diffusion-based methods \cite{jeanneret2022diffusionmodelscounterfactualexplanations, jeanneret2023adversarialcounterfactualvisualexplanations, Jeanneret_2024_WACV, augustin2024digindiffusionguidanceinvestigating, sobieski2024globalcounterfactualdirections, farid2023latentdiffusioncounterfactualexplanations}. Most diffusion-based approaches adapt classifier guidance \cite{dhariwal2021diffusionmodelsbeatgans} to steer the generative process of counterfactuals, requiring access to the classifier and test-time optimization to produce counterfactual images. However, generating counterfactuals this way can be challenging, as the optimization problem closely resembles that of adversarial examples. TIME~\cite{Jeanneret_2024_WACV} proposes an alternative approach by using Textual Inversion~\cite{gal2022imageworthwordpersonalizing} to encode class and dataset contexts into a set of text embeddings, providing a black-box framework for counterfactual explanations. While this removes the need for direct classifier access, Textual Inversion is primarily designed for personalization, focusing on regenerating concepts in novel scenes rather than preserving image structure—an essential aspect of counterfactual generation. 
% Orr: removing due to space and relevancy 
% Other works explore dataset biases by leveraging text-to-image models to induce distributional shifts in data \cite{prabhu2023lancestresstestingvisualmodels, vendrow2023datasetinterfacesdiagnosingmodel}, but these methods are largely constrained to classes that can be easily inverted into text or captions.
% GCD \cite{sobieski2024globalcounterfactualdirections} offers another black-box approach by training a Diffusion Autoencoder \cite{preechakul2022diffusionautoencodersmeaningfuldecodable} alongside a proxy model that approximates the classifier’s objective. However, this method requires training a diffusion model for each dataset, and its proposed sampling strategy may be less effective in complex semantic spaces.
\paragraph{Image Editing.} 
% Recent advances in text-to-image diffusion models \cite{ramesh2022hierarchicaltextconditionalimagegeneration, rombach2022ldm, saharia2022photorealistictexttoimagediffusionmodels} have introduced various semantic control techniques for image editing. 
Recent advances in text-to-image diffusion models \cite{ramesh2022hierarchicaltextconditionalimagegeneration, rombach2022ldm, saharia2022photorealistictexttoimagediffusionmodels, pmlr-v162-nichol22a, flux2024} have enabled test-time controls for image editing, ranging from semantic modifications to attention-based edits and latent space manipulation.
Early approaches applied noise to an image and then denoised it using a new prompt \cite{meng2022sdeditguidedimagesynthesis}, but this often resulted in significant structural changes. Later methods refined direct prompt modifications by incorporating cross-attention manipulations or masking to better preserve image structure \cite{hertz2022prompttopromptimageeditingcross, parmar2023zeroshotimagetoimagetranslation, brack2024leditslimitlessimageediting, 
tumanyan2023plug,
couairon2022diffeditdiffusionbasedsemanticimage}. Brooks \etal \cite{brooks2023instructpix2pixlearningfollowimage} use controlled edits from these methods to train a new diffusion model based on instruction-driven prompts. However, these approaches are limited to text-driven modifications, which restrict the flexibility of edits beyond what can be described with text. Unlike single-image editing methods, Concept Sliders \cite{gandikota2023conceptslidersloraadaptors} introduce a different approach by optimizing a global semantic direction across the diffusion model. While text pairs can guide their optimization, they also propose visual sliders based on image pairs. However, the visual slider approach struggles with unpaired data. 
\paragraph{Diffusion Models with Image Prompts.}
Text-to-image diffusion models generate images from text prompts, but text often falls short in capturing nuanced concepts. Image prompts offer a richer alternative, conveying nuanced details more effectively, as "a picture is worth a thousand words." DALL-E 2~\cite{ramesh2022hierarchicaltextconditionalimagegeneration} pioneered this by conditioning a diffusion decoder on CLIP image embeddings, aided by a diffusion prior for text mapping. Later works offer different architectures~\cite{razzhigaev2023kandinskyimprovedtexttoimagesynthesis} or adapt text-to-image models for image prompts~\cite{ye2023ipadaptertextcompatibleimage,arar2023domainagnostictuningencoderfastpersonalization, flux2024, guo2024pulidpurelightningid}.
\paragraph{Diffusion Inversion.} Editing a real image typically requires first obtaining a latent representation that can be fed into the model for reconstruction. This latent representation can then be modified, either directly or by altering the generative process, to produce the desired edit. Most diffusion-based inversion methods rely on the DDIM \cite{song2022denoisingdiffusionimplicitmodels} sampling scheme, which provides a deterministic mapping from a single noise map to a generated image \cite{mokady2022nulltextinversioneditingreal, wallace2022edictexactdiffusioninversion, parmar2023zeroshotimagetoimagetranslation}. However, this approach introduces small errors at each diffusion step, which can accumulate into significant deviations, particularly when using classifier-free guidance \cite{ho2022classifierfreediffusionguidance}.
Instead of predicting an initial noise map that reconstructs the image through deterministic sampling, an alternative approach considers DDPM \cite{ho2020denoisingdiffusionprobabilisticmodels} sampling and inverts the image into intermediate noise maps \cite{wu2022unifyingdiffusionmodelslatent}. Building on this, Huberman-Spiegelglas \etal \cite{hubermanspiegelglas2024editfriendlyddpmnoise} proposed an inversion technique for the DDPM sampler, along with an edit-friendly noise space better suited for editing applications. In our work, we utilize this technique, while conditioning on image prompts. 
\paragraph{Machine Teaching.} Machine teaching optimizes human learning via computational models. Early work framed this as an optimization task, minimizing example sets for efficient teaching \cite{zhu2015machine}. Generally, the field of machine learning for discovery has machine teaching as a goal \cite{jumper2021highly, chiquier2023muscles}. Recent advances leverage generative models and LLMs for cross-modal discovery, synthesizing representations for conceptual learning \cite{chiquier2024evolving}, decoding structures in mathematics, or programs for scientific discovery \cite{mall2025disciple, romera2024mathematical}. Parallel efforts amplify subtle signals for perception: language models detect fine-grained textual differences \cite{dunlap2024vibecheck}, while video motion magnification enhances visual cues \cite{liu2005motion, wu2012eulerian, oh2018learning}. These methods, though effective for fine-grained discrimination, typically require aligned, abundant data and focus on single modalities. Our work extends these efforts, using diffusion models to generate visual counterfactuals for nuanced category learning.

%% file: sec_arxiv/3_method.tex
\section{Method}
\label{sec:method}
\begin{figure*}[t]
  \centering
  \includegraphics[width=\linewidth]{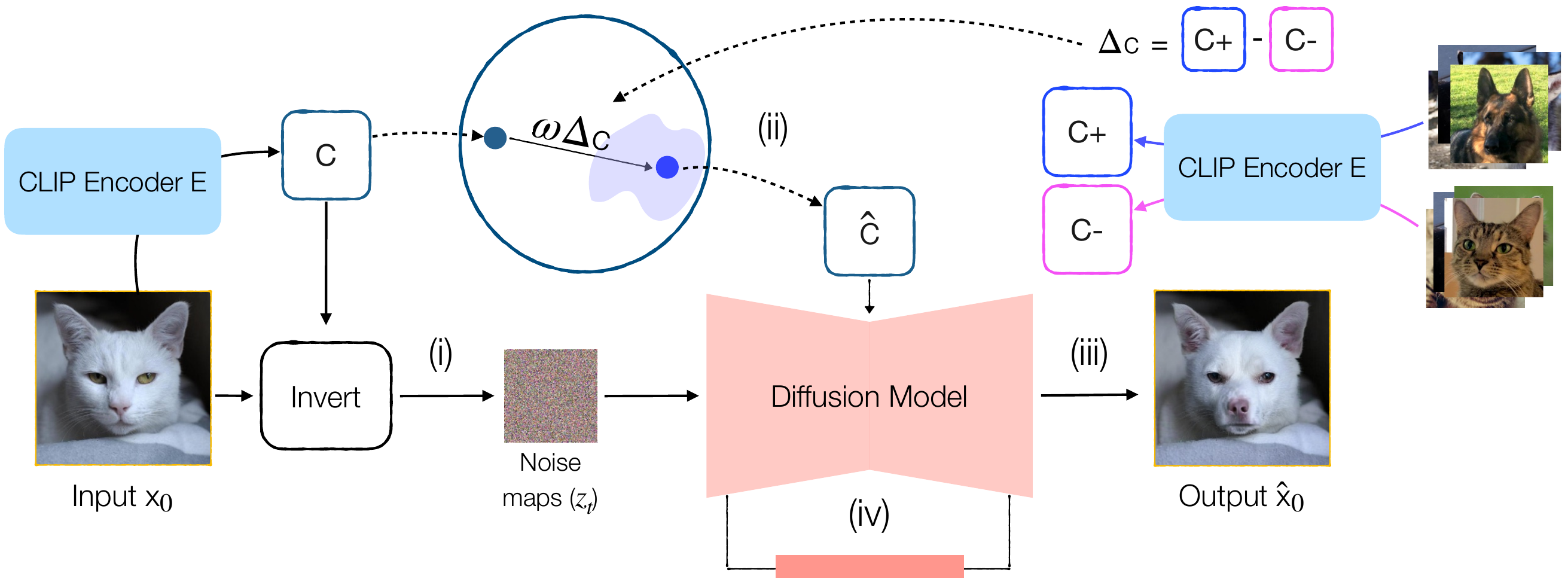} 

   \caption{\textbf{\textit{DIFF}usion method.} Our method consists of four parts. (i) Inverting the real image with DDPM-EF to obtain noise maps. (ii) Performing conditioning space arithmetic using positive and negative embeddings obtained from the training set. (iii) Generation via diffusion sampling, starting from the inverted noise conditioning on the manipulated conditioning vector $\hat{c}$. (iv) Optional domain tuning, in which we fine-tune the diffusion model for domain adaptation. }
   \label{fig:method-diagram}
\end{figure*}

We begin by introducing \textit{DIFF}usion for counterfactual image generation, as illustrated in \Cref{fig:method-diagram}. In \Cref{subsec: preliminaries}, we provide the necessary background on diffusion models. In \Cref{subsec:DIFFusion}, we present our proposed method, outlining its design and implementation.

\subsection{Diffusion Preliminaries}
\label{subsec: preliminaries}

Diffusion models generate data by sampling from a distribution through iterative denoising of noisy intermediate vectors. A forward process is first applied, where noise is gradually added to a clean image $x_0$ over $T$ steps. A noisy sample at timestep $t$ can be expressed as
\begin{equation}
    x_t = \sqrt{\Bar{\alpha}_t} x_0 + \sqrt{1-\Bar{\alpha}_t} \epsilon , \quad t = 1,...,T
    \label{eq:forward-diffusion}
\end{equation}
where $\epsilon \sim \mathcal{N}(0,\textbf{I})$, $\alpha_t$ is a predetermined variance schedule, and $\Bar{\alpha}_t = \prod_{i=1}^T \alpha_i$.  
The model learns to reverse the forward noising process, which can be expressed as an update step over $x_t$,
\begin{equation}
    x_{t-1} = \mu_\theta (x_t, c) + \sigma_t z_t , \quad t = T,...,1
    \label{eq:reverse-diffusion}
\end{equation}
where $z_t$ are i.i.d standard normal vectors, $\sigma_t$ is a variance schedule, and $\mu_\theta (x_t, c)$ is typically parameterized as:
\begin{equation}
    \mu_\theta(x_t, c) = \frac{1}{\sqrt{\alpha_t}} \left( x_t - \frac{1-\alpha_t}{\sqrt{1-\Bar{\alpha}_t}} \epsilon_\theta (x_t, t, c) \right)
    \label{eq:mu}
\end{equation}
Here $\epsilon_\theta(x_t, t, c)$ is the trained noise prediction network, and $c$ is an optional conditioning context, such as an image prompt embedding.

\begin{figure*}[t]
  \centering
  \includegraphics[width=1.0\linewidth]{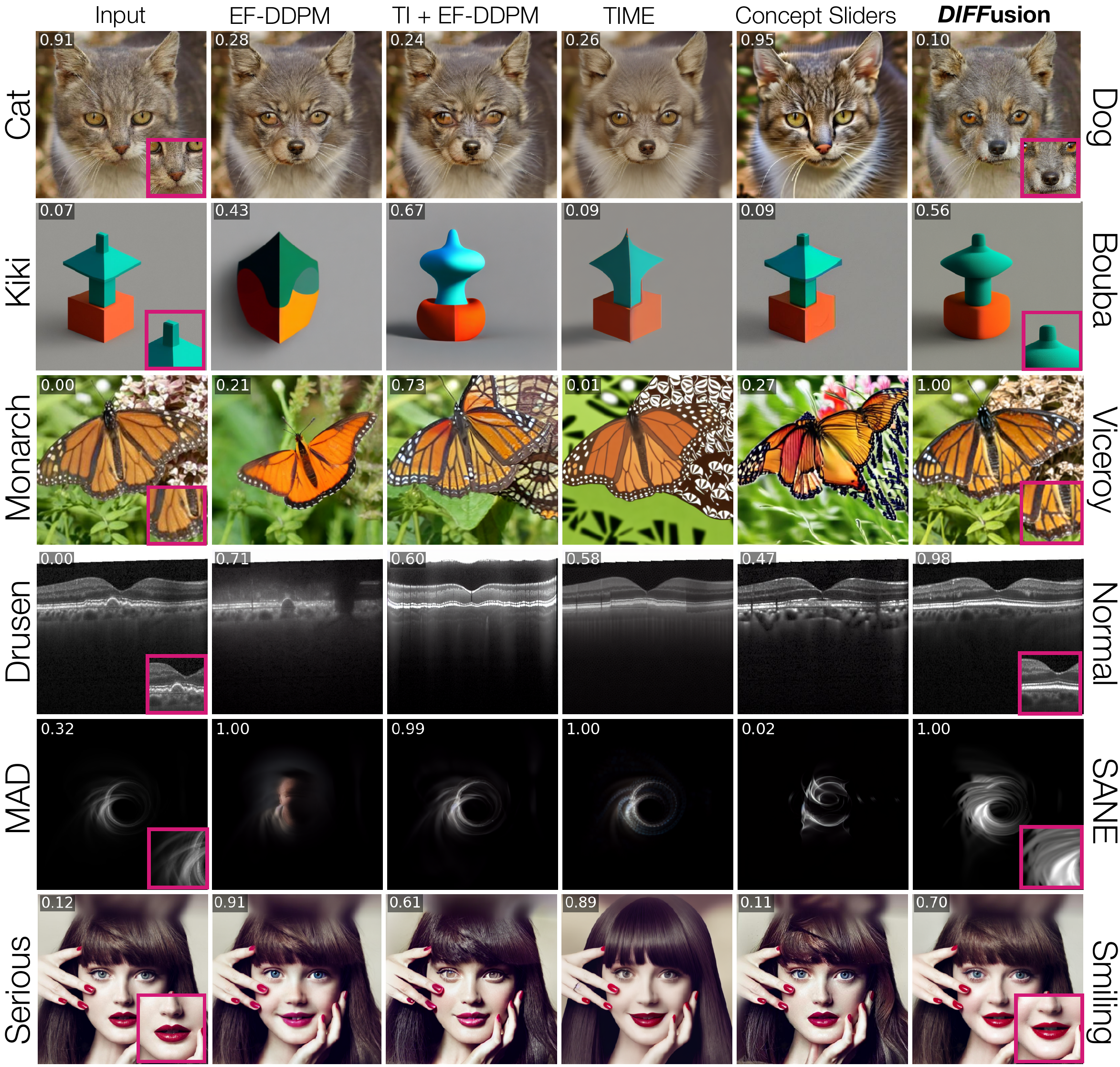}
  \caption{\textbf{Qualitative Results}. We present our qualitative results, where each row corresponds to one direction of our binary datasets. The first column contains the inputs, and each subsequent column contains the results from each baseline, with the last column containing the result from \textbf{\textit{DIFF}usion}. The value in the top left corner of the image is the average probability predicted by our ensemble classifiers. In particular, the magnified boxes in the magenta frame show that our method is able to pick up on small discriminative cues. When converting from MAD to SANE, the whisps become amplified and more uniform in brightness. When converting from Drusen to Normal, the small bumps along the cross-section are flattened out. When converting from Monarch to Viceroy, a cross-sectional line is added on the wing.}
  \label{fig:main_grid}
\end{figure*}

\subsection{\textbf\textit{DIFF}usion}
\label{subsec:DIFFusion}
Given an input image $x_0$, our goal is to find a fine-grained, discriminative edit that changes a classifier's prediction.  
Let $\mathcal{R}_\theta(\mathbf{z}, c)$ be the recursive application of the denoising diffusion model from \Cref{eq:reverse-diffusion}. Our approach finds these edits by inverting the image $x_0$, into a sequence of noise maps,  $\mathbf{z}$, and manipulating the CLIP embeddings of the original image, $c = E(x)$, into a resulting conditioning vector $\hat{c}$, before sampling the modified image. We generate the modified image $\hat{x}_0$ through: 
\begin{align}
\hat{x}_0 = \mathcal{R}_\theta(\mathbf{z}, \hat{c})
\end{align}
Since the diffusion model must generate an image consistent with the original noise maps $\mathbf{z}$, and has a conditioning vector $\hat{c}$ that steers from the source towards the target class, the resulting samples maintain the identity of the original image, but with subtle modifications such that the class label flips. 
\paragraph{Inversion.}
We are interested in extracting noise vectors $\mathbf{z}$, such that, if used in \Cref{eq:reverse-diffusion}, would recover the original image $x_0$. Note that any sequence of $T+1$ images $x_0, ..., x_T$ can be used to extract consistent noise maps for reconstruction by isolating $z_t$ from \Cref{eq:reverse-diffusion} as
\begin{equation}
    z_t = \frac{x_{t-1} - \mu_\theta(x_t, c)}{\sigma_t}, \quad t = T,...,1
    \label{eq:noise-maps}
\end{equation}
We follow the choice suggested in \cite{hubermanspiegelglas2024editfriendlyddpmnoise}
and compute the noise maps through the standard forward diffusion process \Cref{eq:forward-diffusion}, but using statistically independently-sampled noise for each timestep. This yields noise maps $\mathbf{z} = \{x_T, z_T,\dots,z_1\}$ that are consistent with $x_0$. 

\paragraph{Conditioning.}
We generate edits that flip the category through arithmetic operations on $c$, resulting in $\hat{c}$. We apply an additive translation to the conditioning vector $c=E(x)$:
\begin{equation}
\label{eq:manip}
    \hat{c} = c + \omega \Delta c   
\end{equation}
where $c$ is the CLIP image embedding of the original image, $\Delta c$ is a direction that moves the class from the original class to the target class, and $\omega$ is a scaler that varies the direction's strength. We calculate this translation through the difference of means for each class:
\begin{align}
\Delta c = \mathbb{E}_{x_p}\left[ E(x_p) \right] - \mathbb{E}_{x_n} \left[ E(x_n) \right]
\end{align}
such that $x_p$ is an image of class $p$ and $x_n$ is an image of class $n$ (e.g, positive and negative classes). We normalize all the image embeddings with L2 norm prior to the arithmetic.
\paragraph{Sampling.}
We use $\hat{c}$ as the conditioning vector for DDPM sampling, paired with the inverted noise maps, $z$, to generate the counterfactual image. As suggested in \cite{hubermanspiegelglas2024editfriendlyddpmnoise}, we run the generation process starting from timestep $T - T_{skip}$, where $T_{skip}$ is a parameter controlling the resemblance to the input image. Therefore, similar to \Cref{eq:reverse-diffusion}, denoting the denoised edited image at timestep $t$ as $\hat{x}_t$ we have,
\begin{equation}
    \hat{x}_{t-1} = \mu_\theta (\hat{x}_t, \hat{c}) + \sigma_t z_t , \quad t = T - T_{skip},...,1
    \label{eq:edit-reverse-diffusion}
\end{equation}
This approach allows us to systematically steer the image generation toward the target class by adjusting the manipulation scale $\omega$, while maintaining key structural features of the original image through $T_{skip}$.
Intuitively, a larger $T_{skip}$ results in fewer denoising steps under the manipulated condition $\hat{c}$, leading to greater adherence to the input image.
\paragraph{Domain Tuning}
We use a pre-trained diffusion model \cite{kandinsky2.2} that conditions on CLIP image embeddings. When adapting to a new domain, we fine-tune the model using LoRA \cite{hu2021loralowrankadaptationlarge}, training only its cross-attention and corresponding projection layers. As discussed in \Cref{sec:supp-add-lora-free}, we find that domain tuning is beneficial for the Butterfly \cite{van2018inaturalist} and Retina \cite{kermany2018identifying} datasets, but has minimal impact on the other datasets.

%% file: sec_arxiv/4_experiments.tex
\begin{table}[h!]
    \centering
    \begin{tabular}{ll}
        \toprule
        \textbf{Dataset} & \textbf{Class 0 / Class 1} \\
        \midrule
        AFHQ \cite{AFHQ}        & Dog / Cat \\
        KikiBouba \cite{alper2024kikiboubasoundsymbolism}           & Kiki / Bouba \\
        Retina \cite{kermany2018identifying}           & Drusen / Normal \\
        Black-Holes       & MAD / SANE \\
        Butterfly \cite{van2018inaturalist}     & Monarch / Viceroy \\
        CelebA-HQ \cite{CelebAHQ}     & Smile / No-Smile \\
        \bottomrule
    \end{tabular}
    \caption{Datasets and their classification tasks.}
    \label{tab:datasets}
\end{table}

\begin{table*}[t!]
\small
  \caption{We report the accuracy per dataset, for our method and baselines. SR = Success Ratio, LPIPS = Perceptual Distance. In \textbf{bold} are the best results, and in \textit{italic} are the second best results.}
  \label{tab:results}
  \begin{tabular}{@{}llllllllllllllll@{}}
    \toprule
    Method & \multicolumn{2}{c}{AFHQ} & \multicolumn{2}{c}{KikiBouba} & \multicolumn{2}{c}{Retina} & \multicolumn{2}{c}{Black-Holes} & \multicolumn{2}{c}{Butterfly} & \multicolumn{2}{c}{Celeba-HQ-Smile}\\
    \cmidrule(lr){2-3} \cmidrule(lr){4-5} \cmidrule(lr){6-7} \cmidrule(lr){8-9} \cmidrule(lr){10-11} \cmidrule(lr){12-13}
    & SR $\uparrow$ & LPIPS $\downarrow$ & SR $\uparrow$ & LPIPS $\downarrow$ & SR $\uparrow$ & LPIPS $\downarrow$ & SR $\uparrow$ & LPIPS $\downarrow$ & SR $\uparrow$ & LPIPS $\downarrow$ & SR $\uparrow$ & LPIPS $\downarrow$ \\
    \midrule
    EF-DDPM & \textbf{1.0} & \textbf{0.187} & 0.68 & 0.343 & 0.39 & 0.272 & \underline{0.73} & 0.117 & 0.86 & 0.328 & \textbf{1.0} & \textbf{0.104} \\
    TI + EF-DDPM & \textbf{1.0} & \underline{0.211} & \underline{0.97} & 0.332 & \underline{0.89} & 0.330 & 0.5 & \textbf{0.045} & \textbf{1.0} & \underline{0.289} & \textbf{1.0} & 0.181 \\
    TIME & 0.95 & 0.217 & 0.17 & \textbf{0.170} & 0.50 & 0.358 & 0.52 & 0.086 & 0.13 & 0.320 & 0.79 & 0.166 \\
    Concept Sliders & 0.49 & 0.375  & 0.13 & 0.206 & 0.48 & \underline{0.248} & 0.53 & 0.155 & 0.27 & 0.362 & 0.21 & 0.238 \\
    \textbf{\textit{DIFF}usion} & \textbf{1.0} & 0.245  & \textbf{0.98} & \underline{0.176} & \textbf{0.98} & \textbf{0.217} & \textbf{1.0} & \underline{0.076} & \textbf{1.0} & \textbf{0.218} & \textbf{1.0} & \underline{0.116} \\ 
    \bottomrule
  
  \end{tabular}
  
\end{table*}

\section{Experiments}
\label{sec:experiments}

\subsection{Datasets and Baselines}
\label{subsec:datasets-baselines}

\textbf{Datasets}. We quantitatively benchmark on datasets from diverse domains. We also note the corresponding directions under examination for each dataset. We evaluate on AFHQ \cite{AFHQ}, CelebaHQ \cite{CelebAHQ} and KikiBouba \cite{alper2024kikiboubasoundsymbolism} as our non-scientific datasets. We also evaluate on three scientific datasets. The first is Retina \cite{kermany2018identifying}, a dataset of retina cross-sections, both diseased and healthy. The second is Black Holes, which is a dataset of images taken from fluid simulations of accretion flows around a black hole \cite{wong2022patoka}. The simulations assume general relativistic magnetohydrodynamics (GRMHD) under one of two regimes: magnetically arrested (MAD) or standard and normal evolution (SANE) \cite{jiang2023two}. Finally, we also evaluate on Monarch and Viceroy, a fine-grained species classification task. Monarch butterflies evolved to be mimics of Viceroys, and the two species are notoriously difficult to tell apart. 
\newline
\newline
\textbf{Baselines.} We use TIME~\cite{Jeanneret_2024_WACV} as our counterfactual baseline, and replace black-box classifier labels with ground truth labels. For editing baselines, we compare against Stable Diffusion~\cite{rombach2022ldm} with EF-DDPM inversion~\cite{hubermanspiegelglas2024editfriendlyddpmnoise} using class-name prompts. To better accommodate visual concepts, we implemented another baseline that uses Textual Inversion ~\cite{gal2022imageworthwordpersonalizing} for each class of images and then applies source and target prompts based on the desired edit direction. We term this baseline TI + EF-DDPM. Lastly, we use the visual sliders objective of Concept Sliders ~\cite{gandikota2023conceptslidersloraadaptors} that provides a visual counterpart to text-driven attribute edits. To ensure a robust evaluation, we experimented with varying the rank and number of images used for defining the concept direction, selecting the best configuration for each dataset. Since the original method assumes paired data, we adapted it for unpaired settings.

% Finally, Concept Sliders~\cite{gandikota2023conceptslidersloraadaptors} provide visual edits, adapted for unpaired data with optimized settings per dataset. 

\subsection{Editing Results}
\label{subsec: editing-results}

We quantitatively evaluate how well our method can make minimal edits to the image to flip the classifier's prediction. Since our method can generate different strengths of edits, to pick the minimal edit, we generate 10 edits with varying strengths using the $T_{skip}$ parameter, as does the TIME baseline ~\cite{Jeanneret_2024_WACV}, testing from highest to lowest $T_{skip}$, and select the first edit that flips the classifier prediction while maximizing LPIPS similarity to the original image.
% We introduce a counterfactual generation method that minimally edits images to flip classifier predictions. By manipulating two key parameters, tskip (temporal skipping) and manip scale, we control the edit's strength. Our methodology systematically explores 10 tskip values across three manip scales to find the most subtle edit that flips the classifier's prediction.
% For each manip scale, we select the tskip value that successfully changes the prediction while minimizing the LPIPS distance from the original image. When no tskip level induces a prediction change, we default to tskip = 0.1, representing the maximal possible edit. We then pick the best manip scale per dataset. This approach ensures we identify the most minimal intervention necessary to change the classifier's prediction. 
\newline
\newline
\textbf{Metrics.} We evaluate our method using two key metrics. Success Ratio (SR): Also known as Flip-Rate, quantifies the ability of a method to flip an oracle classifier's decision. The oracle classifier we use is an ensemble of ResNet-18 \cite{he2015deepresiduallearningimage}, MobileNet-V2 \cite{sandler2019mobilenetv2invertedresidualslinear}, and EfficientNet-B0 \cite{tan2020efficientnetrethinkingmodelscaling}, trained on each dataset. LPIPS \cite{zhang2018perceptual}: Measures the perceptual similarity between the input and generated image, by capturing feature-level difference in a learned embedding space.
\begin{figure}[b]
  \centering
  \includegraphics[width=1.0\linewidth]{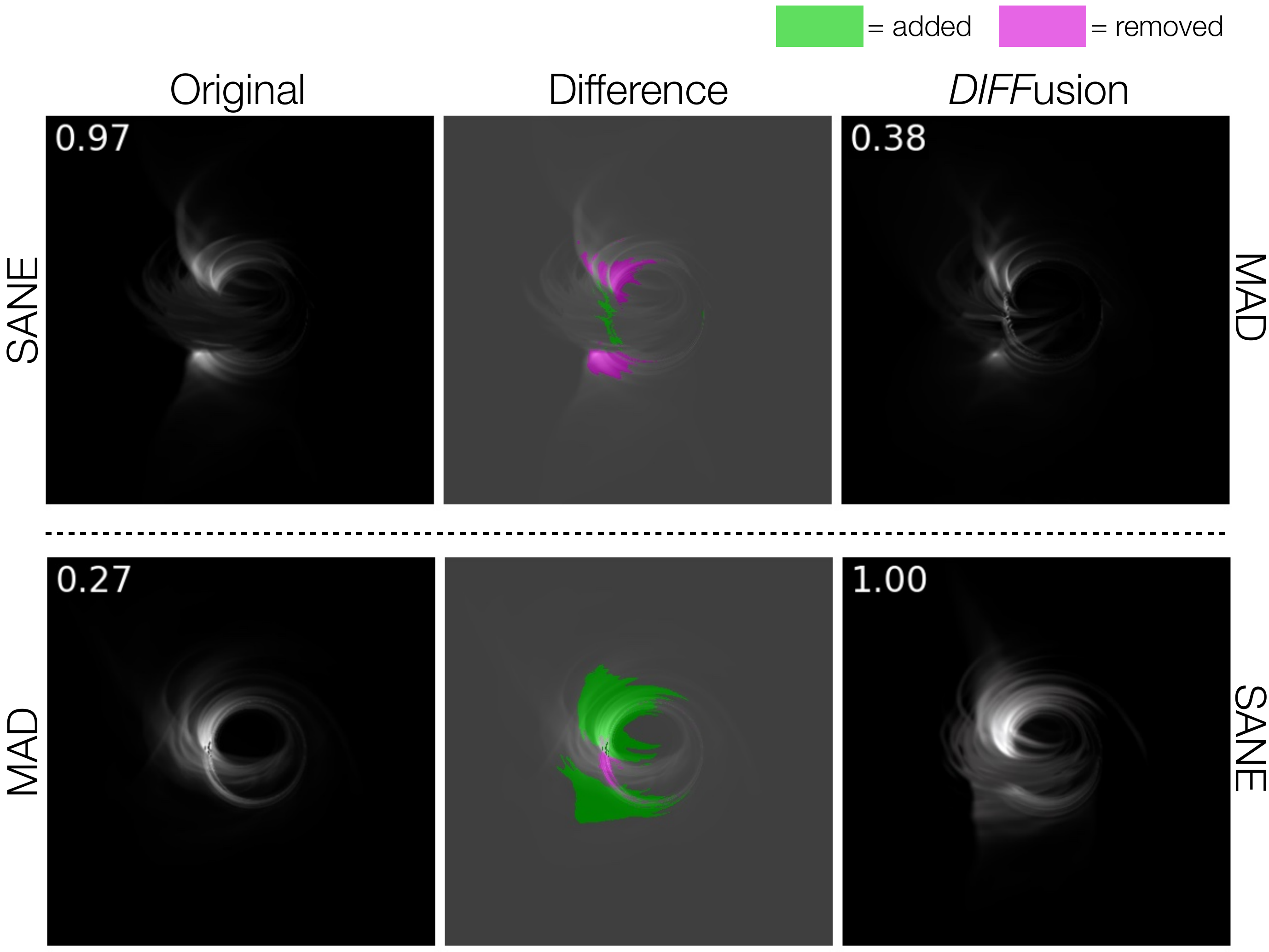}
\caption{\textbf{Original vs. Counterfactual Overlay}. We visualize the difference between the input image and the counterfactual from \textit{DIFF}usion. From SANE to MAD we notice a highlighting of the photon ring (\textbf{\color{green}{green}}). From MAD to SANE we notice that the ring becomes less pronounced (\textbf{\color{magenta}{magenta}}), and wisps appear (\textbf{\color{green}{green}}).}
  \label{fig:overlay}
\end{figure}
% We evaluate ours results with success ratio \cite{Jeanneret_2024_WACV}, and LPIPS \cite{zhang2018perceptual}. \textit{DIFF}usion achieves an SR of at least 0.98 across all datasets, outpacing baselines like TIME (0.17 on KikiBouba) and Concept Sliders (0.21 on Celeba-HQ-Smile). \newline 
% \newline 
% \textbf{Fidelity}: Among methods with SR > 0.50, ours has the lowest LPIPS on KikiBouba (0.176), Retina (0.217), Black-Holes (0.076), and Butterfly (0.218), though it’s edged out slightly on AFHQ (0.245 vs. 0.187) and Celeba-HQ-Smile (0.116 vs. 0.104).\newline 
% \newline 
% \textbf{Balanced Performance}: Unlike EF-DDPM and TI + EF-DDPM, which hit SR 1.0 but sacrifice fidelity (e.g., LPIPS 0.343 and 0.332 on KikiBouba), or TIME, which gets low LPIPS (0.170) but poor SR (0.17), \textit{DIFF}usion excels at both.\newline 
% \newline 
% \textbf{Standout Cases}: On Retina and Black-Holes, we near-perfect SR (0.98 and 1.0) with strong LPIPS (0.217 and 0.076), showing robustness for complex tasks. \newline 
% \newline 
% \textbf{}
% Our results highlight \textit{DIFF}usion’s consistency in balancing classification success and visual quality across diverse datasets. \newline 
\newline
\newline
\textbf{Quantitative Results.} As seen in Table 2, our method achieves the highest SR across all datasets compared to baseline approaches. In terms of LPIPS, it shows significant improvements over previous methods on datasets where language struggles to capture visual details (e.g., Retina, Black-Holes, KikiBouba), unlike datasets with common objects like AFHQ or CelebA-HQ. It also performs either best or competitively on the remaining natural-image datasets. Additionally, while TI + EF-DDPM improves the same text-based baseline, it still struggles with images that are hard to describe textually, such as Black-Holes.
\newline
\newline 
\textbf{Qualitative Results.} In \cref{fig:main_grid}, we present class transitions for all baselines and \textit{DIFF}usion. On familiar datasets like CelebAHQ and AFHQ, our method performs well, similar to baselines. However, its strengths stand out in datasets where language may not fully capture visual details. For KikiBouba, only our method and TI + EF-DDPM round Kiki’s edges, though the baseline changes the original colors, while ours keeps them intact. In the Butterfly dataset, the baselines miss the cross-sectional line, and in the Retina dataset, only our approach removes Drusen while preserving image identity. For the Black-Holes dataset, our method flips the classifier’s prediction with visual features matching SANE’s, even if other methods achieve similar flips. These results suggest our method handles subtle visual nuances particularly well.
\subsection{Teaching Results}
\begin{figure}[b!]
  \centering
  \includegraphics[width=\linewidth]{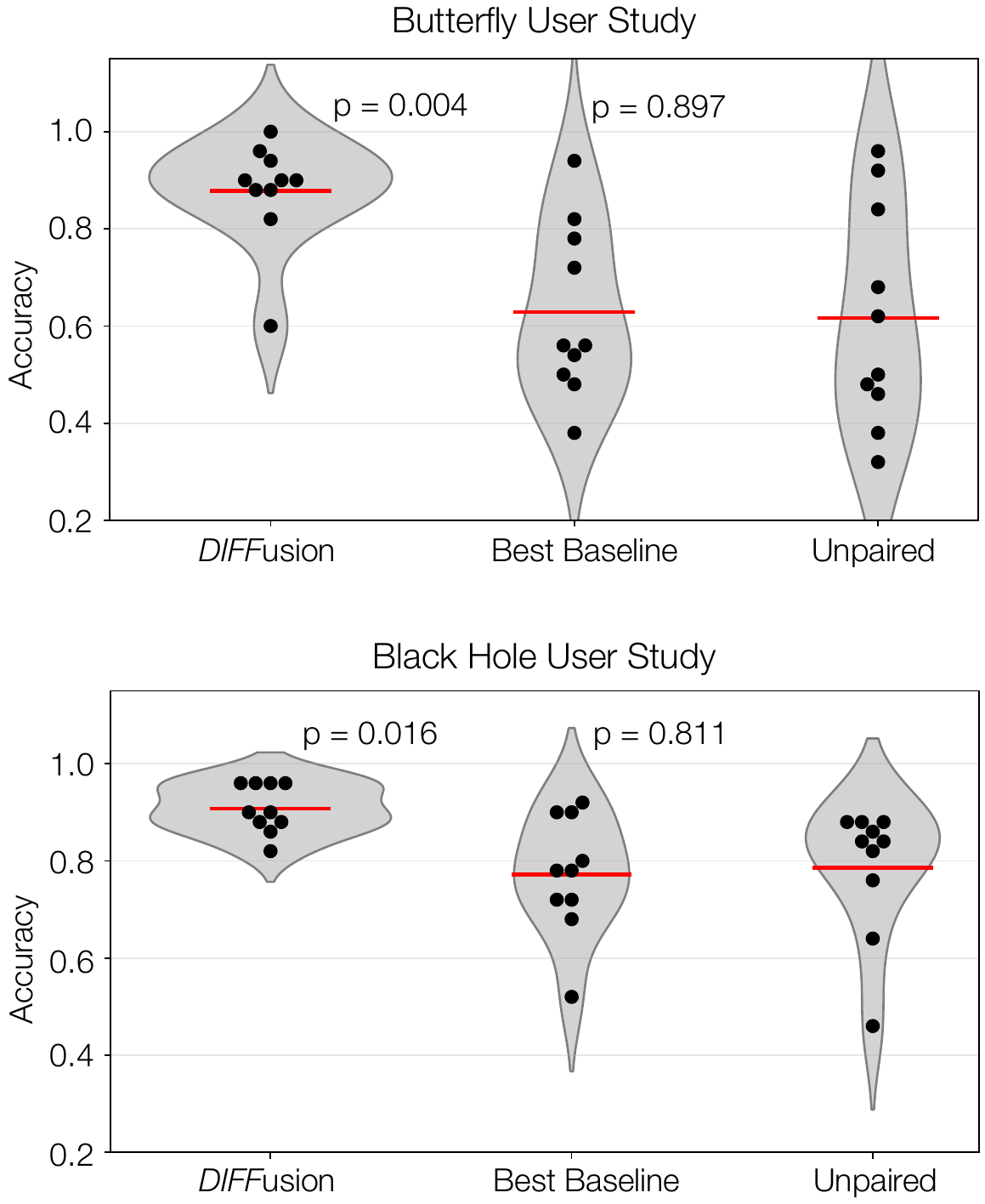}
  \caption{\textbf{User Study Results}. We plot the results from user studies across users who studied our counterfactuals, users who studied the best baseline counterfactuals, and users who studied unpaired images. For both Butterfly and Black Hole datasets, we observe that the users who studied our counterfactuals significantly outperformed the other two groups. The violin plots illustrate the distribution of user percentages, where the width of each grey shape represents the density of data points at corresponding percentages. }
  \label{fig:user_study}
\end{figure}

In this section we wish to evaluate how well our method helps people learn to distinguish subtle visual differences between classes.
\paragraph{User Study Design.}
We divided participants into three groups of 10 people each. Group 1 studied only unpaired images. Group 2 studied videos transitioning from original images to counterfactual images generated by the best baseline. Group 3 studied videos transitioning from original images to counterfactual images generated by our method. Since Groups 2 and 3 viewed transitions from real to edited images, they were also exposed to the unpaired image distribution seen by Group 1. All participants studied their respective materials for 3 minutes to learn to distinguish between the two classes before taking a test. The test required labeling 50 images, evenly distributed with 25 images from each class.
\paragraph{User Study Results.} We assess \textit{DIFF}usion for teaching via a user study on the Black Holes and Butterfly datasets~\cite{van2018inaturalist}, shown in \Cref{fig:user_study}. For Black Holes, unpaired material gave a solid 78\% average score, but our counterfactuals boosted this to 90\%, with 40\% of users hitting near-perfect scores (96\%+), surpassing baselines and counterfactuals. For Butterfly, unpaired data led to varied scores, but our counterfactuals raised 9 out of 10 users above 80\%, standardizing understanding effectively. P-tests confirm significance: Black Holes ($p$ = 0.016 vs. 0.811 for baseline) and Butterfly ($p = 0.004$ vs. $0.897$ for baseline), both $p < 0.05$. Our counterfactuals consistently outperform alternatives, demonstrating the usefulness of our method for teaching humans subtle visual differences.

\begin{figure*}[t!]
  \centering
    \includegraphics[width=0.85\linewidth]{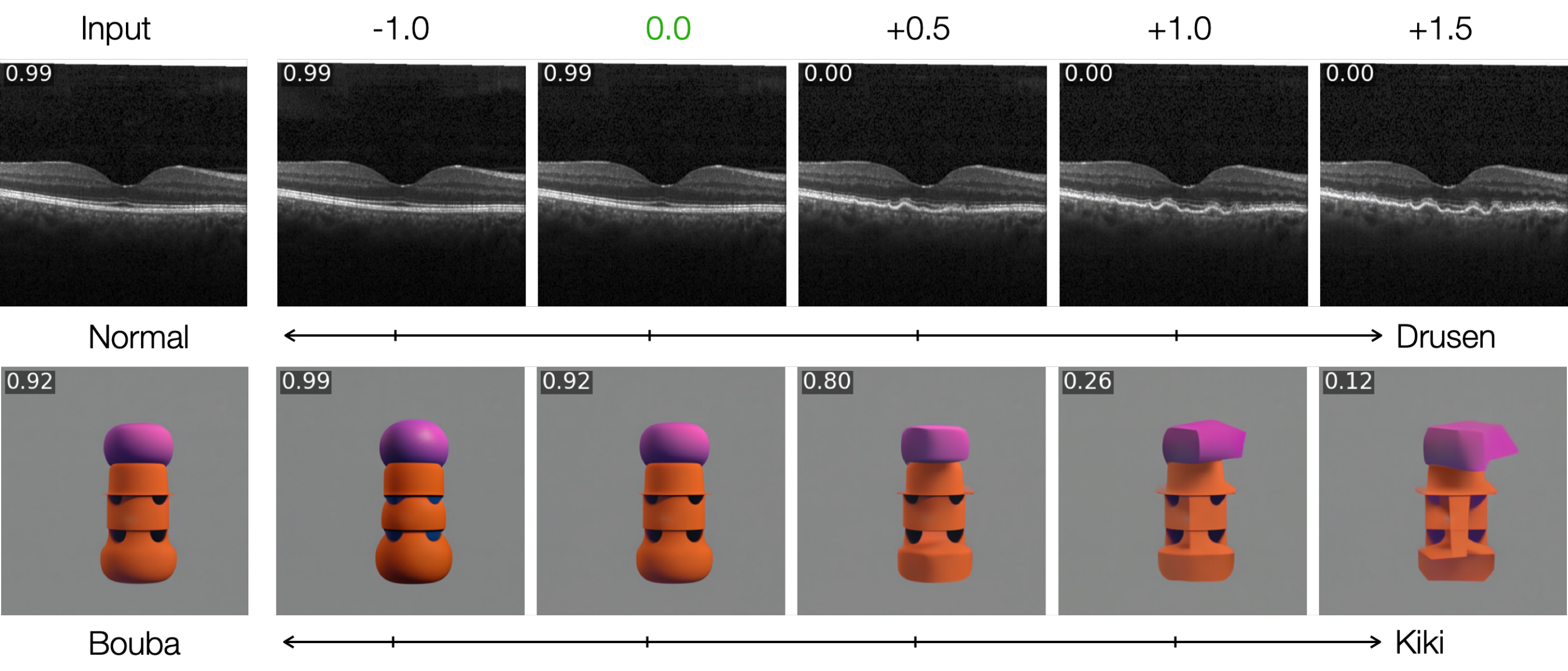}
  \caption{\textbf{Interpolation.} By varying the  manipulation scale $\omega \in \{-0.5, 0.0, 0.5, 1.0, 1.5\}$, we can adjust the manipulation strength, allowing for smooth interpolation between the two classes. Notably, when $\omega=0$ (highlighted in green), we can reconstruct the original image while generally preserving the classifier's probabilities. For a deeper discussion on perfect inversion, refer to \Cref{subsec: supp-perfect-inversion}.}
  \label{fig:interp}
\end{figure*}

\begin{table}[b!]
\centering
\caption{User Study Results - Mean Accuracy (\%)}
\begin{tabular}{@{}l@{\hspace{4pt}}c@{\hspace{4pt}}c@{\hspace{4pt}}c@{}}
\toprule
& Black Holes & Butterfly & Avg. \\
Method & Mean±SD & Mean±SD & Impr. \\
\midrule
Unpaired & 78.6±13.7 & 61.6±22.8 & --- \\
Baseline & 77.2±11.5 & 62.8±16.8 & -0.1\% \\
Ours & 90.8±4.8$$ & 87.8±10.4$$ & +19.2\% \\
\bottomrule
\end{tabular}
\smallskip
\begin{flushleft}
\footnotesize
% *$p < 0.01$ vs. Unpaired\\
% $n = 10$ participants per condition
\end{flushleft}
\label{tab:user_study_results}
\end{table}

\begin{figure}[b!]
  \centering
    \includegraphics[width=\linewidth]{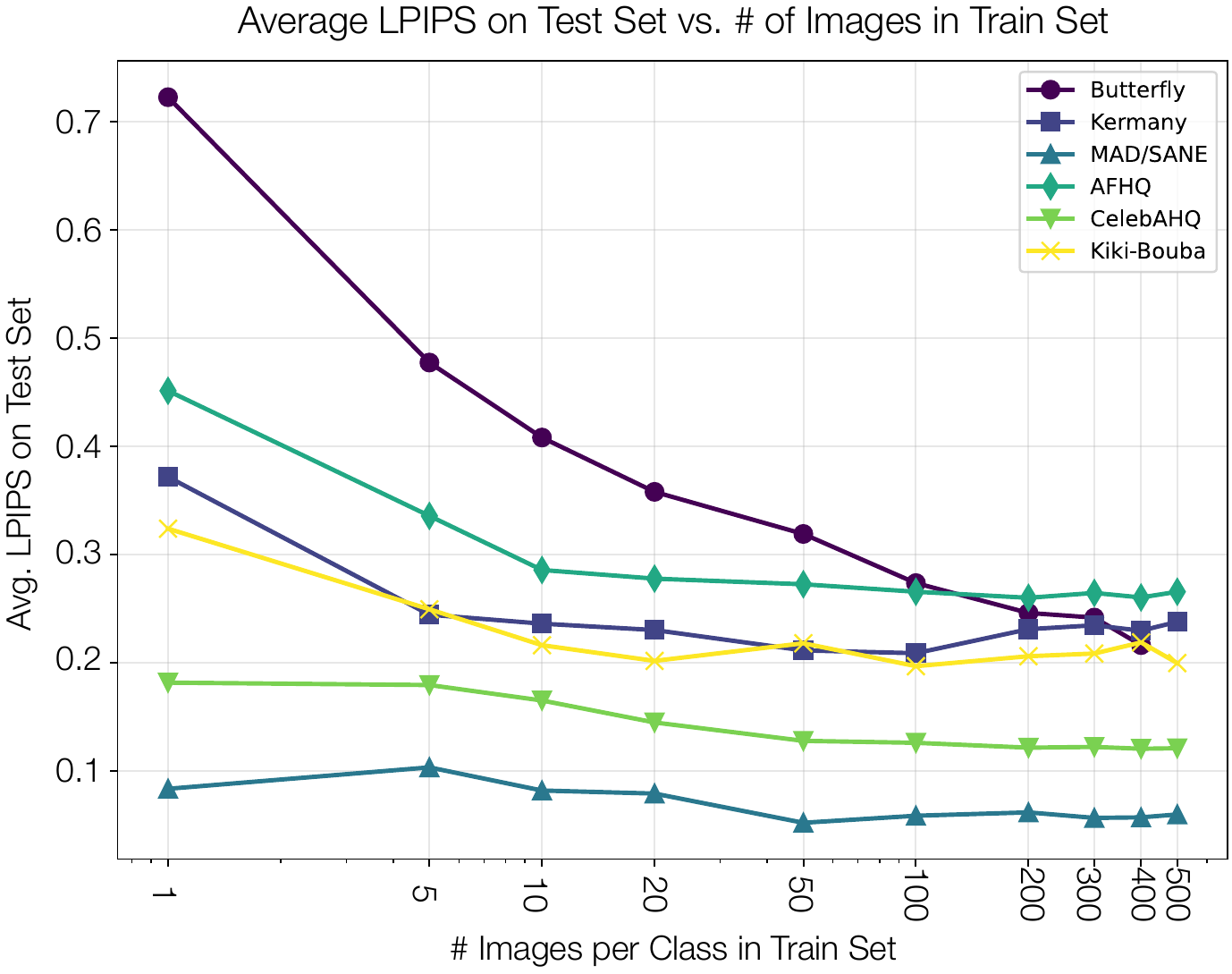}
  \caption{\textbf{Varying num. of Images}. Average LPIPS vs. number of images used per class. LPIPS stabilizes around 50 images for most datasets, reflecting improved identity fidelity and subtle class-distinctive feature shifts with increased embedding samples.}
  \label{fig:numimages}
\end{figure}

% \textbf{Metrics.} We quantify our results with three metrics.  We evaluate leveraging the classifier's accuracy. Since the goal is to edit from one class to another, accuracy measures how often this goal is achieved. We compute the accuracy leveraging both the classifier that was used during guidance, and other classifiers as well. We also report the average change in probabilities. We evaluate with LPIPS \cite{zhang2018unreasonable} to quantify the extent
% of structure preservation (lower is better).\newline 

\subsection{Method Analysis}
\label{subsec:analysis}

\textbf{Varying Dataset Size}. In \cref{fig:numimages}, we examine the impact of varying the number of images per class on the average LPIPS metric across the test sets. We notice that for most datasets, the LPIPS stops improving at around 50 images. In \cref{sec:numimages-grids}, we show qualitative results as the number of images changes. We notice that as the number of images incorporated into the average embeddings increases, the fidelity to the original image’s identity improves, while subtly altering the features that are distinctive between classes. 
\newline
\newline
\textbf{Interpolation}. In \cref{fig:interp}, we present qualitative results demonstrating the effects of varying the manipulation scale, $w$, on an instance of a Normal retina. The manipulation scale, which can take positive or negative values, modulates the transformation direction. Positive values of $w$ shift the features toward Drusen from the Normal retina, while negative values make the image smoother.

\begin{figure}[h!]
  \centering
  \includegraphics[width=\linewidth]{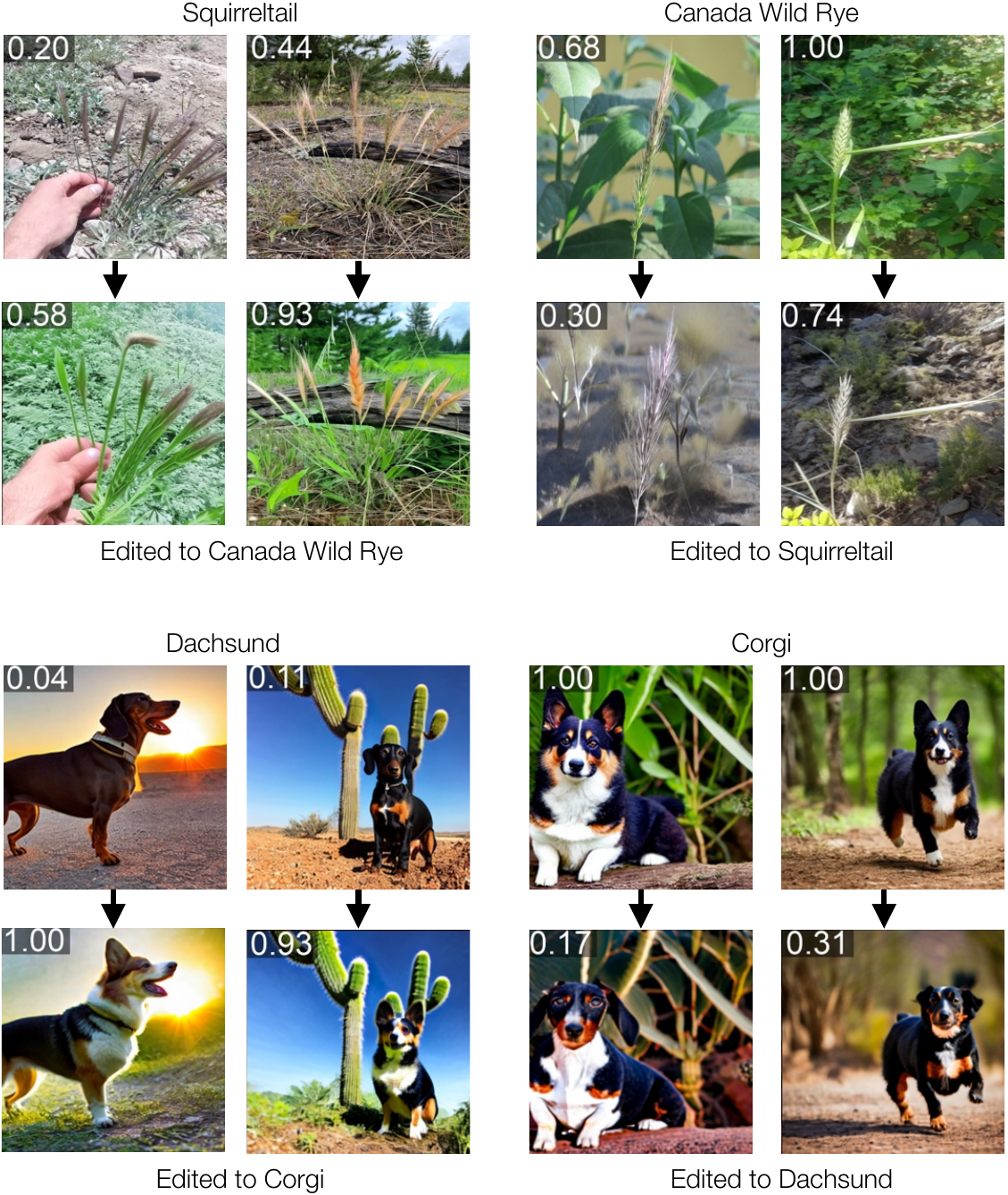}
   \caption{\textbf{Dataset Bias}. \textit{DIFF}usion can reveal dataset bias. Squirreltail-to-Canada Wild Rye shifts emphasize environmental backgrounds over plant traits, reflecting iNaturalist’s contextual bias, while Dachshund-to-Corgi edits prioritize foreground dog features, highlighting variable bias impact.}
   \label{fig:datasetbias}
\end{figure}
% \subsection{Visualizing Dataset Bias} A key insight in our approach is that we edit images by leveraging the statistical differences between two clusters of images—specifically, the mean embeddings of their respective classes. However, this reliance on cluster statistics introduces a sensitivity to dataset bias. If the features that most strongly distinguish the two clusters are not the ones we intend to manipulate—due to biases in the dataset—our edits will reflect those unintended differences instead. This behavior is both a limitation and a strength. As a limitation, it means our method may not always produce the precise edit we aim for. Yet, as a strength, it serves as a tool to visualize and amplify any biases present in the dataset, offering a window into its underlying structure.
\subsection{Visualizing Dataset Bias}
Our method edits images using differences between class mean embeddings, making it sensitive to dataset bias. If distinguishing features reflect unintended biases rather than targeted traits, edits deviate from our intent. This is both a limitation—preventing precise control—and a strength, as it visualizes dataset biases, revealing underlying structure. We show how dataset bias is captured by our method in \cref{fig:datasetbias}. In iNaturalist \cite{van2018inaturalist}, counterfactuals from Squirreltail (dry climates) to Canada Wild Rye (humid) shift backgrounds more than plant structure, suggesting environmental bias dominates. Conversely, using the Spawrious \cite{lynch2023spawriousbenchmarkfinecontrol} dataset, Dachshund-to-Corgi counterfactuals prioritize dog features (e.g., shape, size) over jungle-to-desert backgrounds. We attribute this to stronger foreground differences in dogs and clearer object-background separation, unlike plants blending into settings in iNaturalist data. The effect of dataset bias on edits varies with class prominence and context.

%% file: sec_arxiv/5_discussion.tex
\section{Discussion and Limitations}
\label{sec:discussion}
\textit{DIFF}usion  generates counterfactuals to support visual expertise training across domains with limited data. It reveals dataset biases, often shifting unintended features due to embedding reliance, which limits precise control. Additionally, the arithmetic is very simple: a difference of averages, highlighting a trade-off between flexibility and specificity. Future work could explore disentanglement or guidance mechanisms to enhance edit precision in specialized applications. \newline 
\newline 
\textbf{Acknowledgments:} 
We thank our user study participants and collaborators for their insights. This work is supported by the Carver Mead New Adventures Fund, a Pritzker Award, an AI4Science Amazon Discovery Grant, the NSF AI Institute for Artificial and Natural Intelligence (ARNI), NSF CAREER \#2046910, NSF RETTL \#2202578, DARPA ECOLE, and a Google Fellowship. Views are ours, not necessarily our sponsors'.
\label{sec:discussion}

%% file: sec_arxiv/X_suppl.tex
\clearpage
\setcounter{page}{1}
\maketitlesupplementary

\appendix
\section{Implementation Details}
\label{sec:supp-implement}

\subsection{Method Implementation Details}
\label{subsec:supp-implement-method}
We utilize the diffusion decoder from \cite{kandinsky2.2} and optionally fine-tune LoRA \cite{hu2021loralowrankadaptationlarge} weights on either subsets or the full dataset as discussed in \Cref{subsec:analysis}. For fine-tuning, we set the LoRA rank to 4, the LoRA scaling factor to $\alpha=8$ and use a base learning rate of $0.003$. 
\newline
For inversion, we adapt the edit-friendly DDPM inversion scheme \cite{hubermanspiegelglas2024editfriendlyddpmnoise} to our diffusion decoder. Specifically, we use CFG \cite{ho2022classifierfreediffusionguidance} in both inversion and generation, motivated by \cite{hubermanspiegelglas2024editfriendlyddpmnoise, deutch2024turboedittextbasedimageediting}. We first aim to find guidance scale parameters that achieve perfect reconstruction, and then use these scales for our method. This process is further discussed in \Cref{subsec: supp-perfect-inversion}. 
\newline
To generate counterfactuals, we manipulate the CLIP image space using \Cref{eq:manip}, adjusting the manipulation guidance scale per dataset ($\omega = 1.0$ for AFHQ, $\omega=2.0$ for the rest of the datasets).

\subsection{Baselines Implementation Details}
As described in \Cref{subsec:datasets-baselines} we compare our method against the following baselines: TIME \cite{Jeanneret_2024_WACV}, Stable Diffusion \cite{rombach2022ldm} with EF-DDPM inversion \cite{hubermanspiegelglas2024editfriendlyddpmnoise}, once using class names as text prompts, and once using learned textual embeddings of a group of each class' images through Textual Inversion \cite{gal2022imageworthwordpersonalizing}. Lastly we also compare to Concept Sliders \cite{gandikota2023conceptslidersloraadaptors}. We used the following third-party implementation in this project:
\begin{itemize}
    \item TIME: \cite{Jeanneret_2024_WACV}. Instead of using their classifier's labels to group the data into two classes, we used the ground-truth labels. Additionally, to perform the evaluations, we used our own ensemble classifiers.  \href{https://github.com/guillaumejs2403/TIME}{official implementation}.
    \item EF-DDPM \cite{hubermanspiegelglas2024editfriendlyddpmnoise}: \href{https://github.com/inbarhub/DDPM_inversion}{official implementation}. For the prompt-based baseline, we use class names as prompts of the form "\textit{a photo of a CLASS-NAME}", where CLASS-NAME can be \textit{"cat", "viceroy", "drusen"}, etc. For the scientific datasets we also include an identifier of the form \textit{"Butterfly", "Black Hole", "Retina"}. For the textual-inversion-based baseline (TI + EF-DDPM), we first invert each class of images into a newly added token and save when optimization is done. Then, we use the source-class token as the inversion prompt, and the target-class token as the generation prompt.
    \item Concept Sliders \cite{gandikota2023conceptslidersloraadaptors}: \href{https://github.com/rohitgandikota/sliders}{official implementation}. We built on the official implementation of {official implementation}, utilizing their Visual Concept Sliders and image editing script. While their approach assumes paired data, we found that unpaired data performs well on datasets with well-aligned classes, such as AFHQ. Conversely, for datasets with less alignment, like Butterfly, training with fewer images—ranging from 5 to 20—slightly improved results. To optimize performance, we varied both the number of training images and the LoRA rank for each dataset, evaluating combinations to select the best one per dataset.
\end{itemize}
As discussed in \Cref{subsec: editing-results}, our evaluation algorithm starts by adjusting parameters that have the strongest manipulation effect first (e.g. guidance scales, $T_{skip}$, and manipulation scale), and stops as soon as a flip occurs, logging the generated image along with the parameters that caused the flip. If no flip is achieved, we apply the final set of parameters designed to induce the most significant edit.

\subsection{User Study Details}
Group 1 studied a folder of unpaired images for 3 minutes. This folder contained images of both classes, with the label of the class written on top of the image. Group 2 studied the unpaired image folder for 1 minute, the counterfactuals generated by the best baseline from Class 0 to Class 1 for another minute, and finally the counterfactuals generated by the best baseline from Class 1 to class 0 for a minute. Group 3 followed the same protocol as Group 2, except with our counterfactuals instead. For both Butterfly and Black hole, the best baseline was TI + DDPM-EF, according to LPIPS. We only showed counterfactuals where the class flipped. 

The participants of the user studies were undergraduates and graduates who volunteered in exchange for baked goods. The supplementary material contains videos of the study material for Group 3, for both the black hole and the butterfly dataset. No user had any prior knowledge about either of the datasets before studying the material and taking the test. 

\section{Additional Experiments}
\label{sec:supp-add}

\subsection{SR vs. LPIPS Curves}
\label{sec:supp-add-curves}

\begin{figure*}[h]
  \centering
  \includegraphics[width=\linewidth]{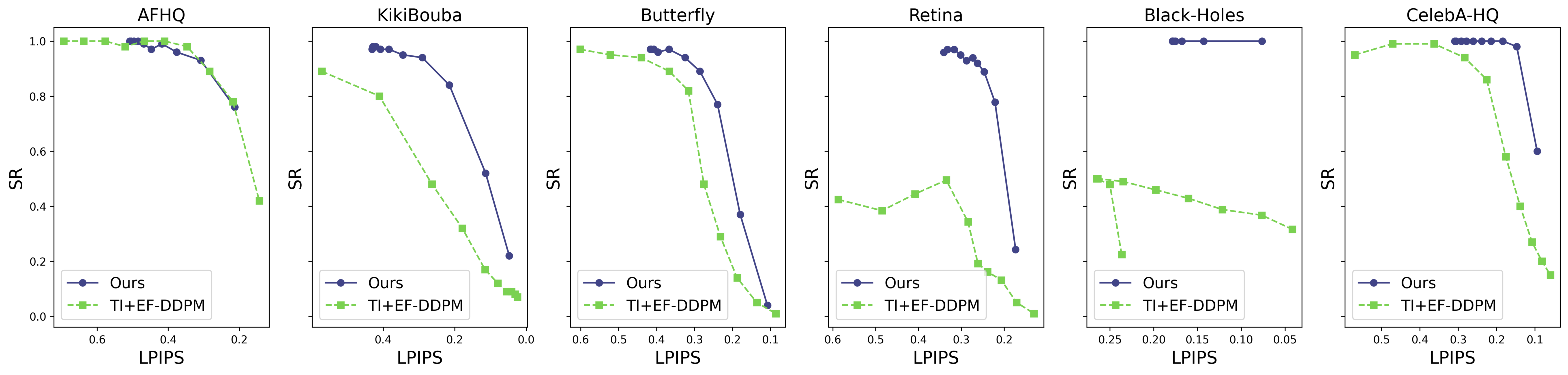}

   \caption{\textbf{Success Ratio (SR) vs. LPIPS curves. } As discussed in \Cref{sec:supp-add-curves}, we fix the guidance scales for each method, and the manipulation scale $\omega$ for ours according to \Cref{sec:supp-implement}. We then vary $T_{skip}$ in increments of 0.1 within the range $[0.0,0.9]$, where $T_{skip}$ represents the percentage of timesteps skipped relative to the total denoising steps.}
   \label{fig:lpips-SR-curve}
\end{figure*}

In \Cref{fig:lpips-SR-curve} we plot the Success Ratio vs. LPIPS curves for our method compared to the best baseline - TI + EF-DDPM, rather than choosing a single set of parameters which is required to report \Cref{tab:results}. Since both use the same inversion technique, we create these curves by varying the $T_{skip}$ parameter. A higher AUC generally indicates a better tradeoff between classifier flip-rate and similarity to input images for each dataset. Our method outperforms the best baseline across all datasets while achieving comparable performance on AFHQ.

\subsection{Results w/o Domain Tuning}
\label{sec:supp-add-lora-free}
As discussed in \Cref{subsec:supp-implement-method}, we begin by fine-tuning a LoRA adapter \cite{hu2021loralowrankadaptationlarge}, applied to the cross-attention and their linear projection weights, using a simple loss at random timesteps $t$, i.e:
\begin{equation*}
    \mathcal{L}_{simple} = \mathbb{E}_{x,\epsilon,t} \left[ \lVert \epsilon - \epsilon_\theta(x_t,t,c) \rVert^2_2 \right]
\end{equation*}
where the prompts $c$ are derived from image embeddings of training set examples. To obtain the best weights, we log SR and LPIPS scores on a small validation set at the end of each epoch. 
In certain datasets, fine-tuning has minimal to no impact on the overall results. This suggests that, in some cases, the prior learned by the pre-trained diffusion model is sufficiently strong to produce meaningful edits when conditioned on a manipulated CLIP image embedding. The results of this analysis are presented in \Cref{tab:lora-free}, where we observe that domain tuning plays a crucial role in datasets like Retina and Butterfly, while having a lesser effect on others.

\subsection{Perfect Inversion}
\label{subsec: supp-perfect-inversion}
\begin{figure*}[b]
  \centering
    \includegraphics[width=0.8\linewidth]{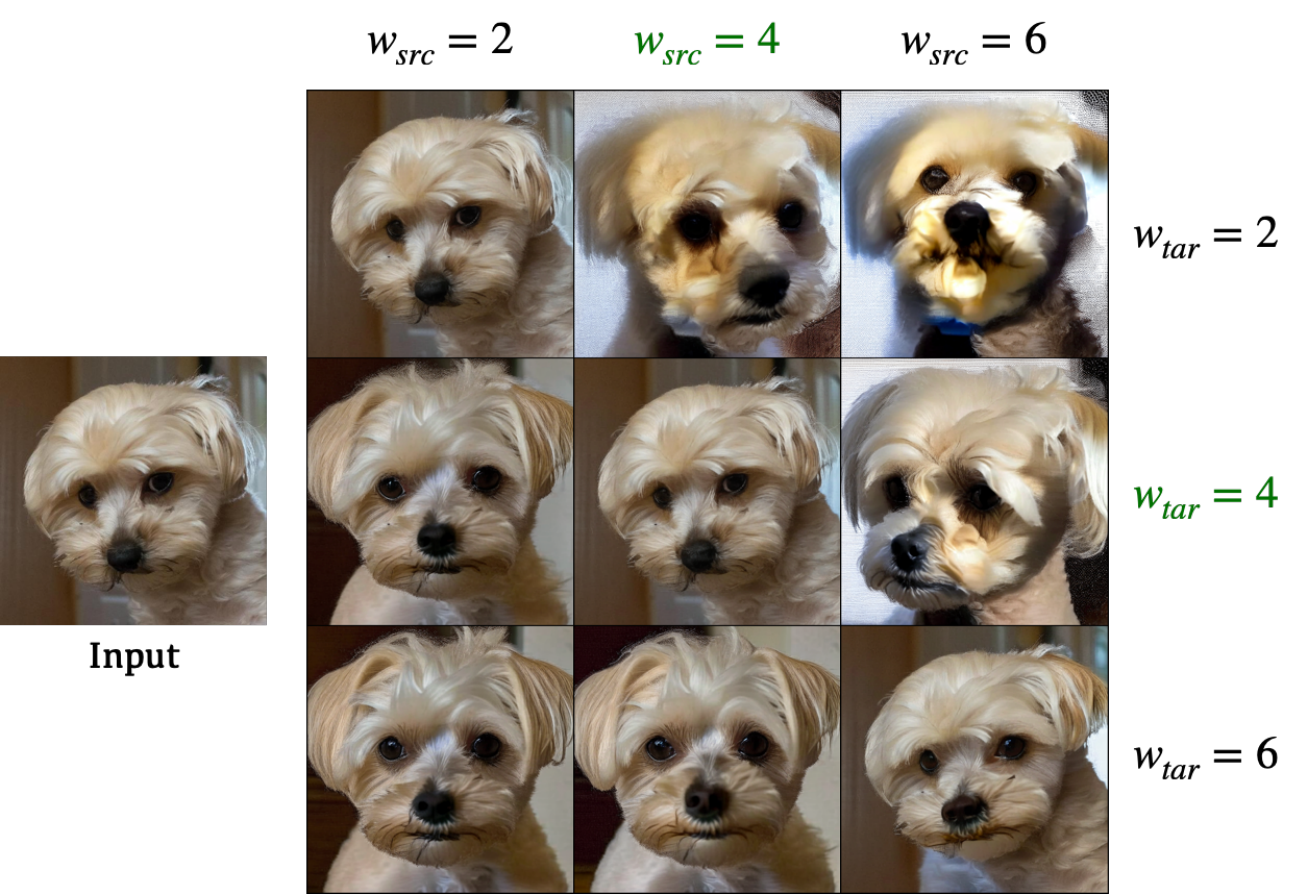}
  \caption{Perfect Inversion.}
  \label{fig:supp-prefect-inversion}
\end{figure*}

Perfect reconstruction can achieved when the same conditioning prompt is used during both inversion and sampling. In this case, we hope that the original image is fully reconstructed. However, DDIM \cite{song2022denoisingdiffusionimplicitmodels} introduces small errors at each timestep, making exact reconstruction challenging, especially with a limited number of timesteps or within the classifier-free guidance framework \cite{ho2022classifierfreediffusionguidance}. Recent works \cite{hubermanspiegelglas2024editfriendlyddpmnoise, wu2022unifyingdiffusionmodelslatent, brack2024leditslimitlessimageediting} focus on non-deterministic DDPM inversion and have demonstrated perfect image reconstruction when applied to Stable Diffusion \cite{rombach2022ldm}. Since we are using an image-conditioned diffusion decoder from the Kandinsky model family \cite{razzhigaev2023kandinskyimprovedtexttoimagesynthesis}, we first explore the choice of guidance scales required to achieve perfect reconstruction while using CFG in both inversion and generation. While perfect reconstruction does not necessarily guarantee a useful editing space, poor reconstruction from the start is likely to cause significant deviations from the source image—an undesirable outcome when generating counterfactuals. \Cref{fig:supp-prefect-inversion} illustrates this effect, showing that using equal guidance terms in inversion and sampling results in good reconstruction, which starts to degrade when inversion guidance scale, $\omega_{src}$, and target guidance scale, $\omega_{tar}$, are larger than 4.

\subsection{Varying Dataset Size Results}
\label{sec:numimages-grids}
\begin{figure*}[t]
  \centering
    \includegraphics[width=0.7\linewidth]{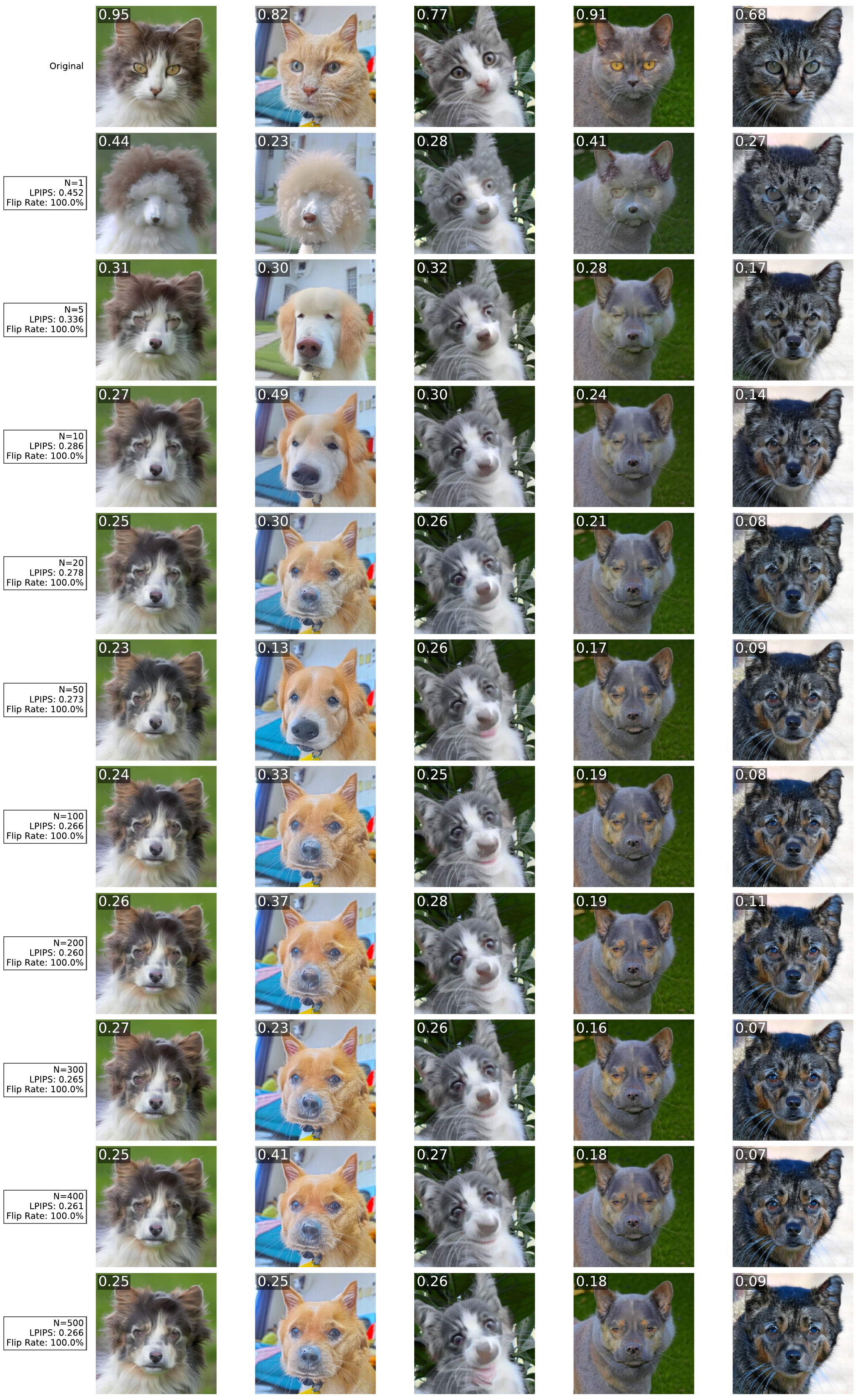}
  \caption{Varying Number of Images for AFHQ.}
  \label{fig:num-images-afhq}
\end{figure*}

\begin{figure*}
  \centering
    \includegraphics[width=0.7\linewidth]{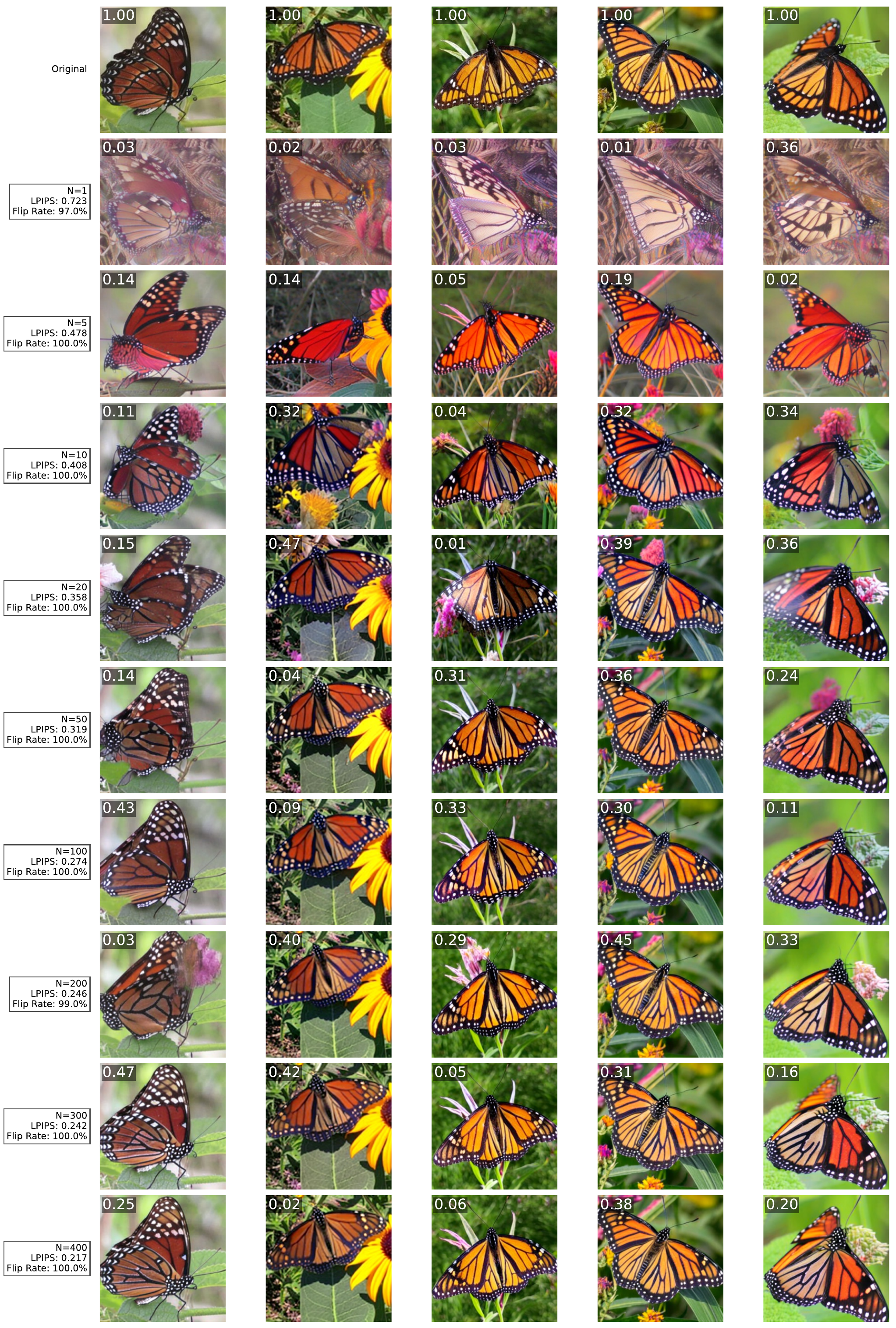}
  \caption{Varying Number of Images for Butterfly.}
  \label{fig:num-images-butterfly}
\end{figure*}

\begin{figure*}
  \centering
    \includegraphics[width=0.7\linewidth]{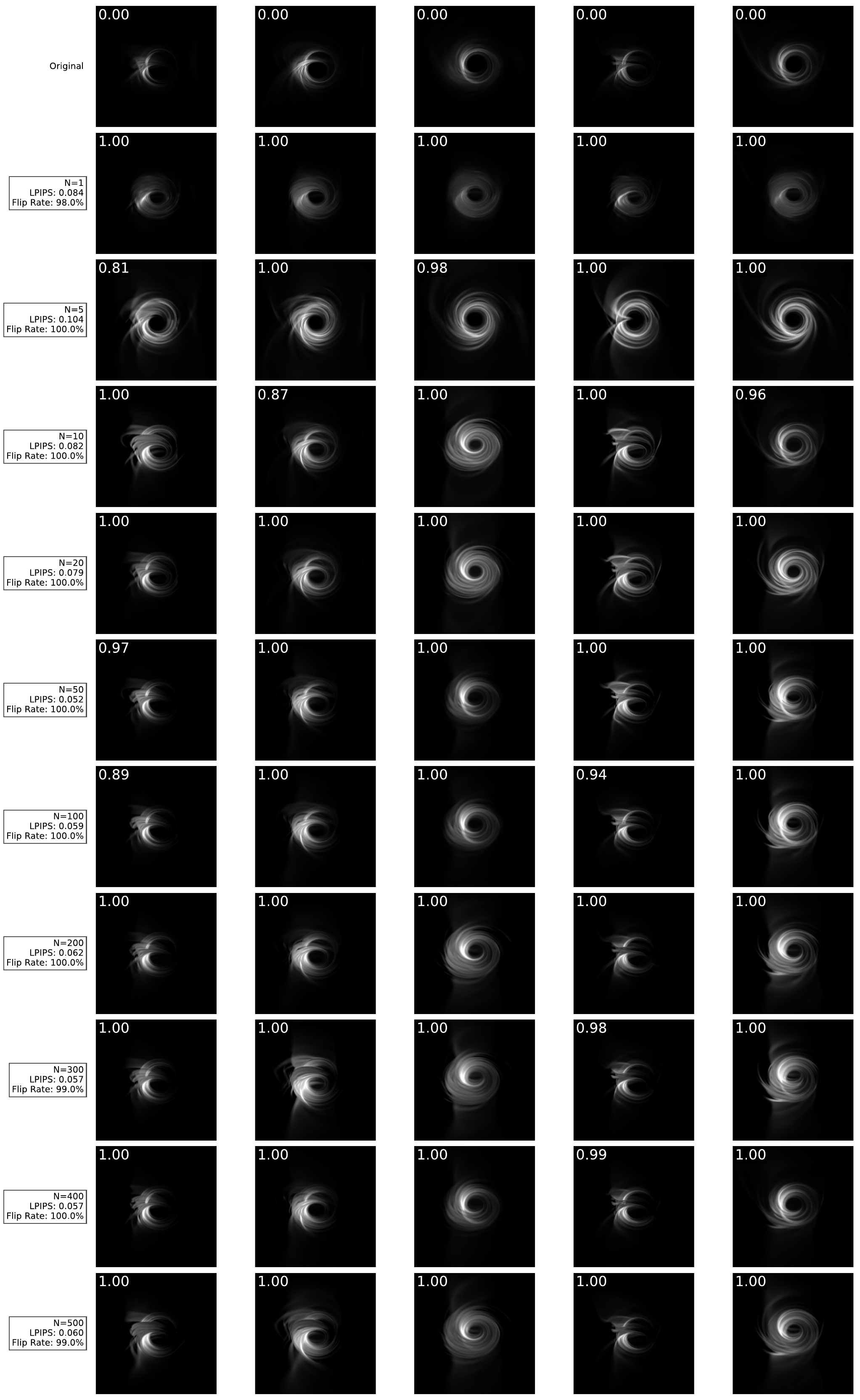}
  \caption{Varying Number of Images for MAD/SANE.}
  \label{fig:num-images-madsane}
\end{figure*}

As described in \Cref{subsec:analysis}, and in \Cref{fig:numimages}, we demonstrate the effect of varying the number of images we have access to for applying \textit{DIFF}usion. In this section, we show examples of generated counterfactuals per number of images in access, $N$, as shown in \Cref{fig:num-images-afhq,fig:num-images-butterfly,fig:num-images-madsane}. We show the grids with increasing number of images so long as the results continue to improve. 

\subsection{Results}
\label{sec:supp-results}

\begin{table*}[t]
\small
  \caption{\textbf{Results without Domain Tuning.} We evaluate our method without fine-tuning on each dataset, measuring performance using Success Ratio (SR) and perceptual distance (LPIPS). Compared to \Cref{tab:results}, we observe substantial improvements in both SR and similarity for the Retina and Butterfly datasets, as well as noticeable gains in reconstruction for Black-Holes and KikiBouba. However, the impact is minimal on the most natural-image datasets, AFHQ and CelebA-HQ. This suggests that the prior learned by the pre-trained diffusion model is strong enough to generate meaningful edits for these datasets without additional fine-tuning.}
  \label{tab:lora-free}
  \begin{tabular}{@{}llllllllllllllll@{}}
    \toprule
    Method & \multicolumn{2}{c}{AFHQ} & \multicolumn{2}{c}{KikiBouba} & \multicolumn{2}{c}{Retina} & \multicolumn{2}{c}{Black-Holes} & \multicolumn{2}{c}{Butterfly} & \multicolumn{2}{c}{Celeba-HQ-Smile}\\
    \cmidrule(lr){2-3} \cmidrule(lr){4-5} \cmidrule(lr){6-7} \cmidrule(lr){8-9} \cmidrule(lr){10-11} \cmidrule(lr){12-13}
    & SR $\uparrow$ & LPIPS $\downarrow$ & SR $\uparrow$ & LPIPS $\downarrow$ & SR $\uparrow$ & LPIPS $\downarrow$ & SR $\uparrow$ & LPIPS $\downarrow$ & SR $\uparrow$ & LPIPS $\downarrow$ & SR $\uparrow$ & LPIPS $\downarrow$ \\
    \midrule
    Ours & 1.0 & 0.249  & 0.98 & 0.2014 & 0.515 & 0.454 & 0.980 & 0.119 & 0.31 & 0.344 & 1.0 & 0.123 \\ 
    \bottomrule
  \end{tabular}
\end{table*}

% \begin{figure*}[t]
%   \centering
%     \includegraphics[width=\linewidth]{figures/num_images_comparison_grid_kikibouba_bouba_to_kiki_lowres.pdf}
%   \caption{Varying Number of Images for KikiBouba}
%   \label{fig:num-images-kikibouba}
% \end{figure*}

% \begin{figure*}
%   \centering
%     \includegraphics[width=\linewidth]{figures/num_images_comparison_grid_kermany_DRUSEN_to_NORMAL_lowres.pdf}
%   \caption{Varying Number of Images for Retina.}
%   \label{fig:num-images-kermany}
% \end{figure*}

% \begin{figure*}
%   \centering
%     \includegraphics[width=0.8\linewidth]{figures/num_images_comparison_grid_celebahq_smiling_to_serious_lowres.pdf}
%   \caption{Perfect Inversion.}
%   \label{fig:num-images-kikibouba}
% \end{figure*}